\documentclass[sigconf]{acmart}
\pdfoutput=1
\usepackage{balance}
\usepackage{booktabs} % For formal tables
\usepackage{soul}
\usepackage{listings}
\usepackage{color}
\definecolor{lightgray}{gray}{0.9}
\usepackage{multirow}
\usepackage{amsmath}
\usepackage{amssymb}
\usepackage{graphicx}
\usepackage[caption=false]{subfig} 
\usepackage{caption}
\usepackage{float}
\usepackage{hyperref}
\usepackage{url}
\usepackage{xspace}
\usepackage{mathrsfs}
\usepackage{bm}
\usepackage{stfloats}
\usepackage{svg}
\usepackage{array}
\usepackage{enumitem}
\usepackage[export]{adjustbox}
\usepackage{amsthm}
\usepackage{algorithm}
\usepackage[noend]{algpseudocode}

\newcommand{\nop}[1]{}

\newcommand{\normal}[1]{#1}

\newcommand{\surfcon}{\textsf{\textsc{SurfCon}}\xspace}

\algnewcommand\algorithmicinput{\textbf{INPUT:}}
\algnewcommand\INPUT{\item[\algorithmicinput]}

\algnewcommand\algorithmicoutput{\textbf{OUTPUT:}}
\algnewcommand\OUTPUT{\item[\algorithmicoutput]}

\algnewcommand\algorithmicphaseone{\textbf{Phase One:}}
\algnewcommand\PHASEONE{\item[\algorithmicphaseone]}

\algnewcommand\algorithmicphasetwo{\textbf{Phase Two:}}
\algnewcommand\PHASETWO{\item[\algorithmicphasetwo]}

\def\BibTeX{{\rm B\kern-.05em{\sc i\kern-.025em b}\kern-.08emT\kern-.1667em\lower.7ex\hbox{E}\kern-.125emX}}
    
\copyrightyear{2019} 
\acmYear{2019} 
\setcopyright{acmcopyright}
\acmConference[KDD '19]{The 25th ACM SIGKDD Conference on Knowledge Discovery and Data Mining}{August 4--8, 2019}{Anchorage, AK, USA}
\acmBooktitle{The 25th ACM SIGKDD Conference on Knowledge Discovery and Data Mining (KDD '19), August 4--8, 2019, Anchorage, AK, USA}
\acmPrice{15.00}
\acmDOI{10.1145/3292500.3330894}
\acmISBN{978-1-4503-6201-6/19/08}

\begin{document}

% The "title" command has an optional parameter, allowing the author to define a "short title" to be used in page headers.
% \title{The Name of the Title is Hope}
\title{
\surfcon: Synonym Discovery on Privacy-Aware Clinical Data
}

% The "author" command and its associated commands are used to define the authors and their affiliations.
% Of note is the shared affiliation of the first two authors, and the "authornote" and "authornotemark" commands
% used to denote shared contribution to the research.
\author{Zhen Wang$^*$, Xiang Yue$^*$, Soheil Moosavinasab$^\dagger$, Yungui Huang$^\dagger$, Simon Lin$^\dagger$, Huan Sun$^*$}
\affiliation{
\institution{$^*$The Ohio State University}
}
\email{{wang.9215, yue.149, sun.397}@osu.edu}
\affiliation{
\institution{$^\dagger$Abigail Wexner Research Institute at Nationwide Children's Hospital}
}
\email{{SeyedSoheil.Moosavinasab, Yungui.Huang, Simon.Lin}@nationwidechildrens.org}

%
% By default, the full list of authors will be used in the page headers. Often, this list is too long, and will overlap
% other information printed in the page headers. This command allows the author to define a more concise list
% of authors' names for this purpose.
\renewcommand{\shortauthors}{Wang, et al.}

%
% The abstract is a short summary of the work to be presented in the article.
\begin{abstract}

Unstructured clinical texts contain rich health-related information. To better utilize the knowledge buried in clinical texts, discovering synonyms for a medical query term has become an important task. Recent automatic synonym discovery methods leveraging raw text information have been developed. However, to preserve patient privacy and security, it is usually quite difficult to get access to large-scale raw clinical texts. In this paper, we study a new setting named \textit{synonym discovery on privacy-aware clinical data} (i.e., medical terms extracted from the clinical texts and their aggregated co-occurrence counts, without raw clinical texts). To solve the problem, we propose a new framework \surfcon that leverages two important types of information in the privacy-aware clinical data, i.e., the \textit{\ul{surf}ace form information}, and the \textit{global \ul{con}text information} for synonym discovery. In particular, the surface form module enables us to detect synonyms that look similar while the global context module plays a complementary role to discover synonyms that are semantically similar but in different surface forms, and both allow us to deal with the OOV query issue (i.e., when the query is not found in the given data). We conduct extensive experiments and case studies on publicly available privacy-aware clinical data, and show that \surfcon can outperform strong baseline methods by large margins under various settings.

\end{abstract}



%
% Keywords. The author(s) should pick words that accurately describe the work being
% presented. Separate the keywords with commas.
\keywords{Synonym Discovery, Privacy-Aware Clinical Data, Medical Term Recommendation}

%
% A "teaser" image appears between the author and affiliation information and the body 
% of the document, and typically spans the page. 
%   \Description{Enjoying the baseball game from the third-base seats. Ichiro Suzuki preparing to bat.}

%
% This command processes the author and affiliation and title information and builds
% the first part of the formatted document.
\maketitle

\section{Introduction}
\label{sec:intro}

Clinical texts in Electronic Medical Records (EMRs) are enriched with valuable information including patient-centered narratives, patient-clinician interactions and disease treatment outcomes, which can be especially helpful for future decision making. To extract knowledge from unstructured clinical texts, synonym discovery \cite{wang2015medical} is an important task which can benefit many downstream applications. For example, when a physician issues a query term (e.g., "vitamin C") to find relevant clinical documents, automatically discovering its synonyms (e.g., "c vitamin", "vit c",  "ascorbic acid") or even commonly misspelled variations (e.g. "viatmin c") can help to expand the query and thereby enhance the retrieval performance. 

\begin{figure}[t]
    \resizebox{\linewidth}{!}{%
    \includegraphics[width=\linewidth, left]{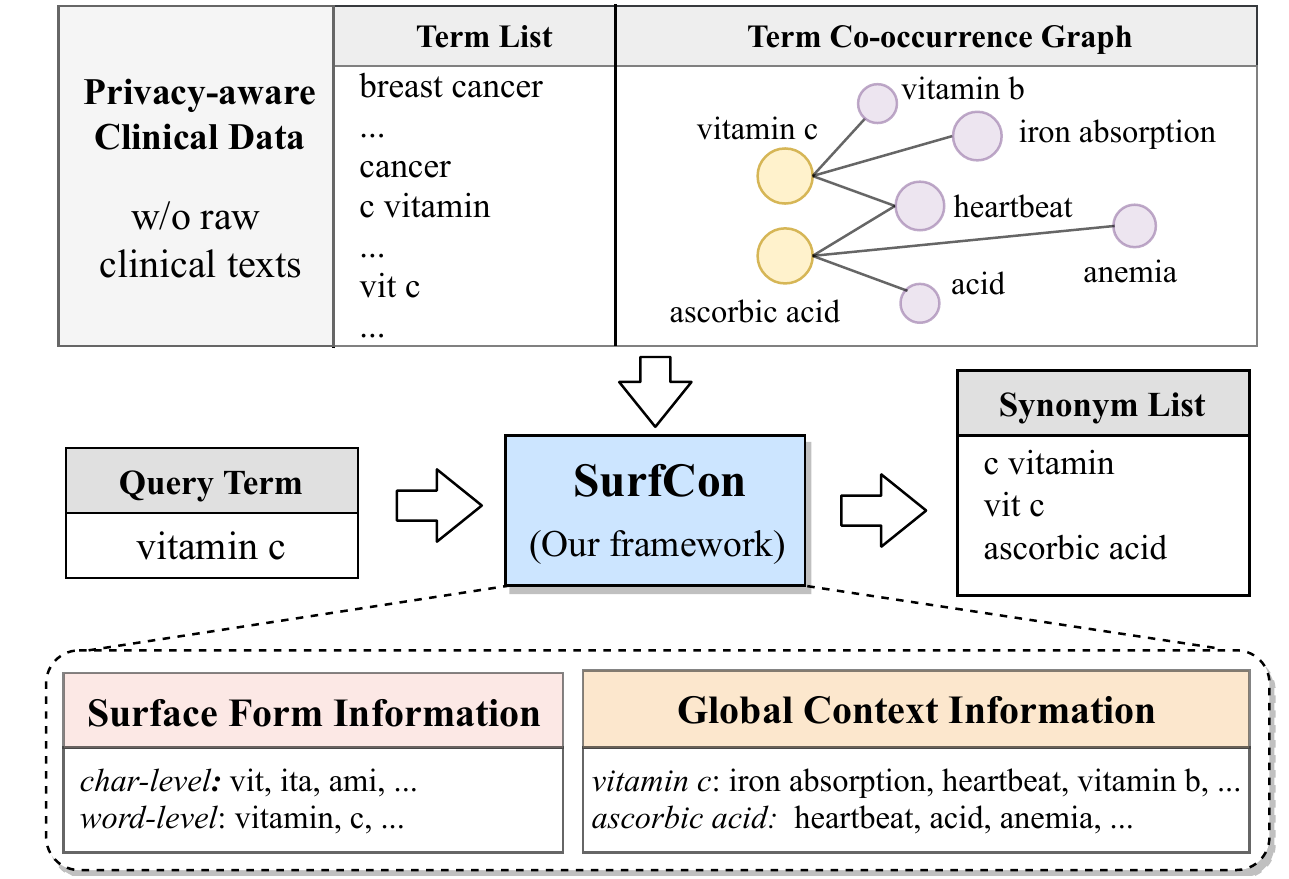}}
    \vspace{-15pt}
    \caption{Task illustration: We aim to discover synonyms for a given query term from privacy-aware clinical data by effectively leveraging two important types of information: Surface form and global contexts. 
    \nop{echo in the introduction}}
    \vspace{-15pt}
    \label{fig:intro_intuition}
\end{figure}

For the sake of patient privacy and security, it is usually quite difficult, if not impossible, for medical institutes to grant public access to large-scale raw or even de-identified clinical texts \cite{beam2018clinical}. Consequently, medical terms\footnote{A medical term is a single- or multi-word string (e.g., "Aspirin", "Acetylsalicylic Acid").} and their aggregated co-occurrence counts extracted from raw clinical texts are becoming a popular (although not perfect) substitute for raw clinical texts for the research community to study EMR data~\cite{finlayson2014building, ta2018columbia, beam2018clinical}.
For example, \citet{finlayson2014building} released millions of medical terms extracted from the clinical texts in Stanford Hospitals and Clinics as well as their global co-occurrence counts, rather than releasing raw sentences/paragraphs/documents from the clinical text corpus. 
In this work, we refer to the given set of medical terms and their co-occurrence statistics in a clinical text corpus as \textit{privacy-aware} clinical data, and {investigate synonym discovery task on such data ({Figure \ref{fig:intro_intuition}}): \textit{Given a set of terms extracted from clinical texts as well as their global co-occurrence graph\footnote{where each node is a medical term and each edge between two nodes is weighted by the number of times that two terms co-occur in a given context window.}, recommend a list of synonyms for a query term}. Developing effective approaches under this setting is particularly meaningful, as they will suggest that one can utilize less sensitive information (i.e., co-occurrence statistics rather than raw sentences in clinical texts) to perform the task well}. 

A straightforward approach to obtain synonyms is to map the query term to a knowledge base (KB) entity and retrieve its synonyms or aliases stored in the KBs. However, it is widely known that KBs are incomplete and outdated, and their coverage of synonyms can be very limited~\cite{wang2015knowledge}. In addition, the informal writing of clinical texts often contain variants of surface forms, layman terms, frequently misspelling words, and locally practiced abbreviations, which should be mined to enrich synonyms in KBs. Recent works~\cite{wang2015medical, qu2017automatic, zhang2019synonymnet} have been focused on automatic synonym discovery from massive text corpora such as Wikipedia articles and PubMed paper abstracts. {When predicting if two terms are synonyms or not, such approaches usually leverage the original sentences (a.k.a. \textit{local} contexts) mentioning them, and hence do not apply or work well under our privacy-aware data setting where such sentences are unavailable.}

{Despite the lack of local contexts, {we observe} two important types of information carried in the privacy-aware data - surface form information and global context information (i.e., co-occurrence statistics).} In this work, we aim to effectively leverage these two types of information for synonym discovery, {as shown in Figure \ref{fig:intro_intuition}}. 

Some recent works~\cite{neculoiu2016learning, mueller2016siamese} model the similarity between terms in the character-level. For example, \citet{mueller2016siamese} learn the similarity between two sequences of characters, which can be applied for discovering synonyms that look alike such as "vit c" and "vitamin c". However, we observe two common phenomena that such approaches cannot address well and would induce false positive and false negative predictions respectively: (1) Some terms are similar in surface form but do not have the same meaning (e.g., "hemostatic" and "homeostasis", where the former means a process stopping bleeding while the latter refers to a constant internal environment in the human body); (2) Some terms have the same meaning but are different in surface form (e.g., "ascorbic acid" and "vitamin c" are the same medicinal product but look different).

On the other hand, given a term co-occurrence graph, various distributional embedding methods such as \cite{pennington2014glove, tang2015line, levy2014neural} have been proposed to learn a {distributional} representation (a.k.a. embedding) for each term based on its \textit{global} contexts (i.e., terms connected to it in the co-occurrence graph). The main idea behind such methods is that two terms should have similar embedding vectors if they share a lot of global contexts. However, we observe that the privacy-aware clinical data tends to be very \textit{noisy} due to the original data processing procedure\footnote{\normal{This tends to be a common issue in many scenarios as raw data has to go through various pre-processing steps for privacy concerns.}}, which presents new challenges for utilizing global contexts to model semantic similarity between terms. For example, \citet{finlayson2014building} prune the edges between two terms co-occurring less than 100 times, which can lead to missing edges between two related terms in the co-occurrence graph. \citet{ta2018columbia} remove all concepts with singleton frequency counts below 10.
Hence, \normal{the noisy nature of the co-occurrence graph makes it less accurate to embed a term based on their original contexts. Moreover, when performing the synonym discovery task, users are very likely to issue a query term that does not appear in the given co-occurrence data. We refer to such query terms as Out-of-Vocabulary (OOV). Unlike In-Vocabulary\footnote{Query terms that appear in the given co-occurrence graph are referred to as In-Vocabulary (InV).} query terms, OOV query terms do not have their global contexts readily available in the given graph, which makes synonym discovery even more challenging}.

In this paper, to address the above challenges and effectively utilize both the \ul{surf}ace form and the global \ul{con}text information in the privacy-aware clinical data, we propose a novel framework named {\surfcon} which consists of a bi-level surface form encoding component and a context matching component, both based on neural models. The bi-level surface form encoding component exploits both character- and word-level information to encode a medical term into a vector. It enables us to compute a surface score of two terms based on their encoding vectors. As mentioned earlier, such surface score works well for detecting synonyms that look similar in surface form. However, it tends to miss synonymous terms that do not look alike. Therefore, we propose the context matching component to model the semantic similarity between terms, which plays a complementary role in synonymy discovery.

Our context matching component first utilizes the bi-level surface form encoding vector for a term to predict its potential global contexts. Using predicted contexts rather than the raw contexts in the given graph enables us to handle OOV query terms and also turns out to be effective for InV query terms. Then we generate a semantic vector for each term by aggregating the semantic features from predicted contexts using two mechanisms - static and dynamic representation mechanism. %\fromT{So messy here.}
Specifically, given term $a$ and term $b$, the dynamic mechanism aims to learn to weigh the importance of individual terms in $a$'s contexts based on their {semantic matching degree} with $b$'s contexts, while the static mechanism assigns equal weights to all terms in one's contexts. The former takes better advantage of individual terms within the contexts and empirically demonstrates superior performance.

Our contributions are summarized in three folds:
\begin{itemize}[leftmargin=*]
    \item We study the task of synonym discovery under a new setting, i.e., on privacy-aware clinical data, where only a set of medical terms and their co-occurrence statistics are given, and local contexts (e.g., sentences mentioning a term in a corpus) are not available. It is a practical setting given the wide concern about patient privacy for access to clinical texts and also presents unique challenges to address for effective synonym discovery.
    
    \item We propose a novel and effective framework named \surfcon that can discover synonyms for both In-Vocabulary (InV) and Out-of-Vocabulary (OOV) query terms. \surfcon considers two complementary types of information {based on neural models} - surface form information and global context information of a term, where the former works well for detecting synonyms that are similar in surface form while the latter can help better find synonyms that do not look alike but are semantically similar. 
    
    \item We conduct extensive experiments on publicly available privacy-aware clinical data and demonstrate the effectiveness of our framework in comparison with various baselines and our own model variants.
\end{itemize}

\vspace{-10pt}
\section{Task Setting}
\label{task-setting}
In this section, we clarify several terminologies used in this paper as well as our problem definition:

\noindent
\textbf{Privacy-aware Clinical Data.} 
Electronic medical records (EMRs) typically contain patient medical information such as discharge summary, treatment, and medical history. In EMRs, a significant amount of clinical information remains under-tapped in the unstructured clinical texts. However, due to privacy concerns, access to raw or even de-identified clinical texts in large quantities is quite limited. Also, traditional de-identification methods, e.g., removing the 18 HIPAA identifiers~\cite{stubbs2015annotating}, require significant manual efforts for the annotation~\cite{dorr2006assessing}. Moreover, there also exists the risk that de-identified data can be attacked and recovered by the re-identification in some cases \cite{garfinkel2015identification}. Thus, to facilitate research on EMRs, an increasingly popular substitute strategy for releasing raw clinical texts is to extract medical terms and their aggregated co-occurrence counts from the corpus \cite{beam2018clinical,ta2018columbia, finlayson2014building}. We refer to such data as privacy-aware clinical data in this paper. Converting raw sentences to co-occurrence data protects privacy as original patient records are very unlikely to be recovered. However, the local context information contained in the raw sentences is also lost, which makes various tasks including synonym discovery more challenging under privacy-aware datasets.

\noindent
\textbf{Medical Term Co-occurrence Graph.}
A medical term-term co-occurrence graph is defined as $G$=$(V, E)$, where $V$ is the set of vertices, each representing a medical term extracted from clinical texts. Each vertex has a surface form string (e.g., "vitamin c", "cancer") which is the spelling of the medical term. $E$ is the set of edges, each weighted by how many times two terms co-occur in a certain context window ({e.g., notes from patient records within 1 day}).

\noindent
\textbf{Medical Term Synonym.} 
Synonyms of a medical term refer to other medical terms that can be used as its alternative names~\cite{qu2017automatic}. For example, "vit c", "c vitamin" and "ascorbic acid" refer to the same medicinal product, while "Alzheimer's disease" and "senile dementia" represent the same disease. In our dataset, the extracted medical terms are mapped to the Unified Medical Language System (UMLS) \cite{bodenreider2004unified} Concept Unique Identifier (CUI) {by \cite{finlayson2014building}}. Different terms mapping to the same UMLS CUI are treated as synonyms for {model training/development/testing}.

\noindent
\textbf{Task Definition.} 
We formally define our task of {synonym discovery on privacy-aware clinical data} as: \textit{Given a medical term co-occurrence graph $G$, for a query term $q$ (which can be either In-Vocabulary or Out-of-Vocabulary), recommend a list of medical terms from $G$ that are likely to be synonyms of $q$.
}

\vspace{-5pt}
\section{\surfcon Framework}
\label{sec:framework}

In this section, we introduce our proposed framework \surfcon for synonym discovery on privacy-aware clinical data. 

\vspace{-5pt}
\subsection{Overview}
\label{subsec:framework-overview}
We observe two important types of information carried in the privacy-aware clinical data: surface form information of a medical term and the global contexts from the given co-occurrence graph. On the one hand, existing approaches \cite{neculoiu2016learning} using character-level features to detect synonyms could work well when synonyms share a high string similarity, but tend to produce false positive predictions (when two terms look similar but are not synonyms, e.g., "hemostatic" and "homeostasis") and false negative predictions (when two terms are synonyms but look very different, e.g., "ascorbic acid" and "vitamin c"). On the other hand, the global contexts of a term under the privacy-aware setting tend to be noisy partly due to the original data pre-processing procedure, which also presents challenges for using them to model the semantic similarity between terms. Thus, a framework that is able to effectively leverage these two types of information needs to be carefully designed.

\begin{figure}[t!]
	\centering
    \resizebox{\linewidth}{!}{%
    \includegraphics[width=\linewidth]{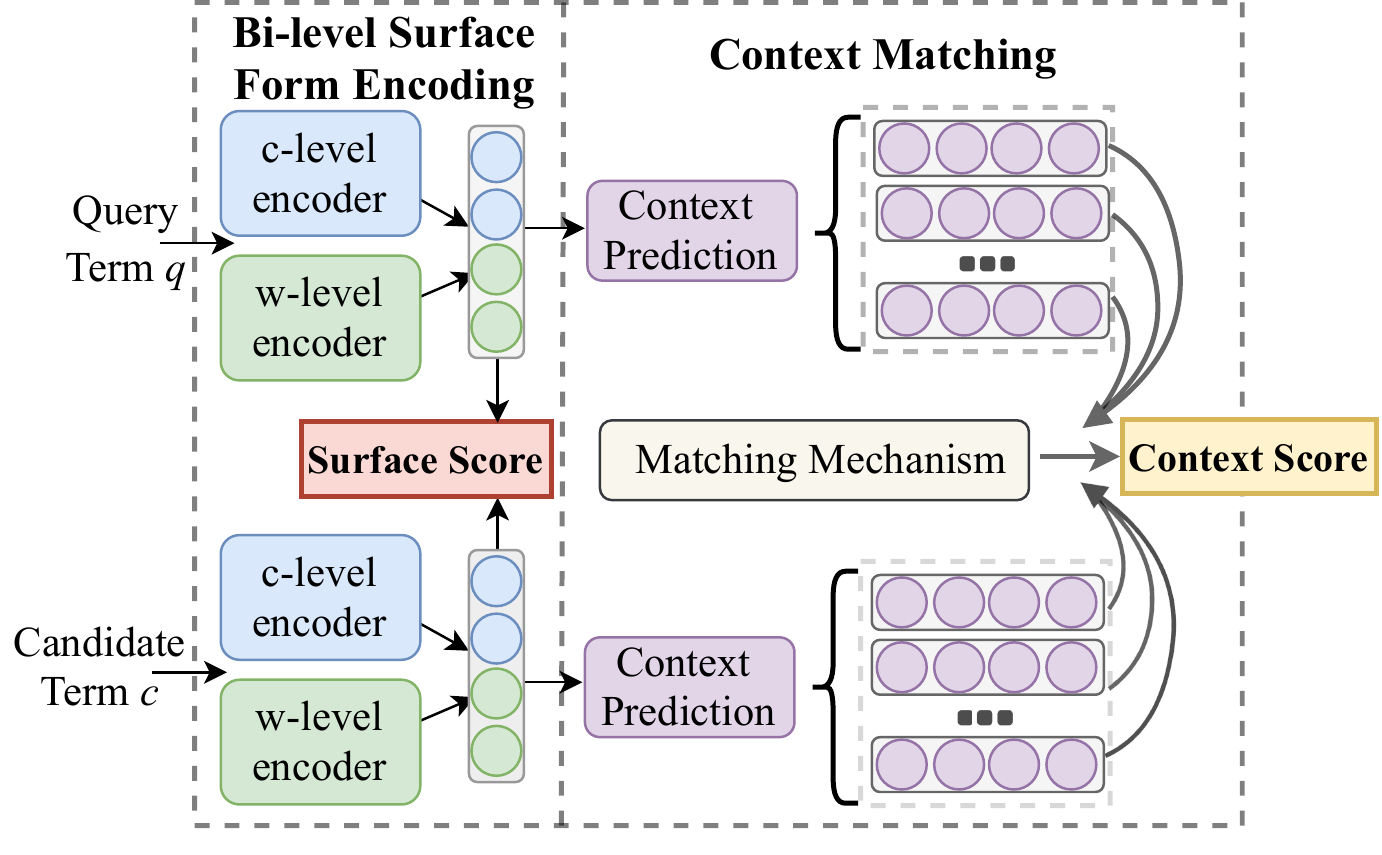}}
    \vspace{-20pt}
	\caption{{Framework overview. For each query term, a list of candidate terms will be ranked based on both the surface and context scores.}}
    \vspace{-15pt}
	\label{fig:framework-overview}
\end{figure}

Towards that end, we propose \surfcon (Figure \ref{fig:framework-overview}) and summarize its high-level ideas as below:

    \noindent
    (1) Given a query term (whether being InV or OOV), the {bi-level surface form encoding component} and the context matching component score a candidate term\footnote{Every term in the given co-occurrence graph can be a candidate term.} respectively based on the surface form information and global context information. The former enables us to find synonyms that look similar to the query term by considering both character- and word-level information, and the latter complements it by capturing the semantic similarity between terms to better address the false positive and false negative problem mentioned earlier.
    
    \noindent 
    (2) Considering the original global contexts being noisy as well as the existence of OOV query terms, instead of directly leveraging the raw global contexts, the context matching component will first utilize the surface form encoding vector of a term to \textit{predict} its potential global contexts\footnote{For terms in the co-occurrence graph, predicting contexts can be treated as denoising its original global contexts (or edges)}. We then investigate a novel dynamic context matching mechanism (see Section \ref{subsubsec:context-matching} for details) to evaluate if two terms are synonyms based on their predicted contexts.

    \noindent
    (3) The two components are combined by a weighted score function, in which parameters are jointly optimized with a widely used ranking algorithm ListNet \cite{cao2007learning}. At testing time, given a query term, candidate terms are ranked based on the optimized score function. 

\vspace{-5pt}
\subsection{Methodology}
Now we describe the two components of \surfcon: Bi-level Surface Form Encoding and Context Matching in details.
\label{subsec:methodology}

\subsubsection{\textbf{Bi-level Surface Form Encoding}}
\label{subsubsec:bi-level-encoding}

The bi-level surface form encoding of our framework aims to model the similarity between two terms at the surface form level, as we observe that two terms tend to be synonymous if they are very similar in surface forms. Such observation is intuitive but works surprisingly well in synonym discovery task. Driven by this observation, we design the bi-level surface form encoding component in a way that both of character- and word-level information of terms are captured. Then, a score function is defined to measure the surface form similarity for a pair of terms based on their bi-level encoding vectors. The bi-level encoders are able to encode surface form information of both InV terms and OOV terms.

Specifically, as shown in Figure \ref{fig:framework-overview}, given a query term $q$ and a candidate term $c$, we denote their character-level sequences as $x_q=\{x_{q, 1}, ..., x_{q, m_q}\}, x_c=\{x_{c, 1}, ..., x_{c, m_c}\}$, and their word-level sequences as $w_q=\{w_{q, 1}, ..., w_{q, n_q}\}, w_c=\{w_{c, 1}, ..., w_{c, n_c}\}$, where $m_q,n_q,m_c,n_c$ are the length of the character-level sequence and word-level sequence of the query term and the candidate term respectively. Then we build two encoders $\text{ENC}^{ch}$ and $\text{ENC}^{wd}$ to capture the surface form information at the character- and word-level respectively:
\begin{equation}
\label{eqn:encoder}
\small
    \begin{aligned}
    s_q^{ch}&=\text{ENC}^{ch}(x_{q,1},...,x_{q,m_q}), 
    s_q^{wd}=\text{ENC}^{wd}(w_{q,1},...,w_{q,n_q})\\
    s_c^{ch}&=\text{ENC}^{ch}(x_{c,1},...,x_{c,m_c}), 
    s_c^{wd}=\text{ENC}^{wd}(w_{c,1},...,w_{c,n_c})
    \end{aligned}
\end{equation}
\noindent
where $s_q^{ch}, s_c^{ch}\in \mathbb{R}^{d_c}$ are the character-level embeddings for the query and candidate terms, and $s_q^{wd},s_c^{wd}\in \mathbb{R}^{d_w}$ are the word-level embeddings for the query and candidate terms respectively.

Note that there has been a surge of effective encoders that model sequential information from character-level or word-level, ranging from simple look-up table (e.g., character n-gram~\cite{hashimoto2017jmt} and Skip-Gram~\cite{mikolov2013distributed}) to complicated neural network architectures (e.g., CNN~\cite{kim2016character}, LSTM~\cite{ballesteros2015improved} and Transformer~\cite{vaswani2017attention}, etc.). For simplicity, here, we adopt simple look-up tables for both character-level embeddings and word-level embeddings. Instead of randomly initializing them, we borrow pre-trained character n-gram embeddings from \citet{hashimoto2017jmt} and word embeddings from \citet{pennington2014glove}. Our experiments also demonstrate that these simple encoders can well encode surface form information of medical terms for synonym discovery task. We leave {evaluating} more complicated encoders as our future work.

After we obtain the embeddings at both levels, we concatenate them and apply a nonlinear function to get the surface vector $s$ for the query and candidate term. Let us denote such encoding process as a function $h(\cdot)$ with the input as term $q$ or $c$ and the output as the surface vector $s_q$ or $s_c$:
\begin{equation}
\begin{aligned}
   s_q&=h(q)=\text{tanh}( [s_q^{ch},s_q^{wd}] W_s + b_s),\\ 
   s_c&=h(c)=\text{tanh}( [s_c^{ch},s_c^{wd}] W_s + b_s)
\end{aligned}
\end{equation}
\noindent
where the surface vectors $s_q, s_c\in \mathbb{R}^{d_s}$, and $W_s \in \mathbb{R}^{(d_c+d_w)\times d_s}, b_s
\in \mathbb{R}^{d_s}$ are weight matrix and bias for a fully-connected layer.

Next, we define the surface score for a query term $q$ and a candidate term $c$ to measure the surface form similarity based on their encoding vectors $s_q$ and $s_c$:
\begin{equation}
    \textsf{Surface Score}\,(q, c) = f_s(s_q, s_c)
\end{equation}

\subsubsection{\textbf{Context Matching}}
\label{subsubsec:context-matching}

In order to discover synonyms that are not similar in surface form, and also observing that two terms tend to be synonyms if their global contexts in the co-occurrence graph are semantically very relevant, we design the context matching component to capture the semantic similarity of two terms by carefully leveraging their global contexts. We first illustrate the intuition behind this component using a toy example:

\newtheorem{exam}{Example}
\begin{exam}
\label{example:contecxt-matching}
\textbf{\emph{[Toy Example for Illustration.]}} Assume we have a query term \textit{"vitamin c"} and a candidate term \textit{"ascorbic acid"}. The former is connected with two terms \textit{"iron absorption"} and \textit{"vitamin b"} in the co-occurrence graph as global contexts, while the latter has \textit{"fatty acids"} and \textit{"anemia"} as global contexts.
\end{exam}
\noindent
Our context matching component essentially aims to use a term's contexts to represent its semantic meaning and a novel \textit{dynamic context matching mechanism} is developed to determine the importance of each individual term in one's contexts. For example, \textit{"iron absorption"} is closely related to \textit{"anemia"} since the disease "anemia" is most likely to be caused by the iron deficiency. Based on the observation, we aim to increase the relative importance of \textit{"iron absorption"} and \textit{"anemia"} in their respective context sets when representing the semantic meaning of \textit{"vitamin c"} and \textit{"ascorbic acid"}. Therefore, we develop a novel dynamic context matching mechanism to be introduced shortly.

In order to recover global contexts for OOV terms and also noticing the noisy nature of the co-occurrence graph mentioned earlier, we propose an \textit{inductive context prediction module} to predict the global contexts for a term based on its surface form information instead of relying on the raw global contexts in the given co-occurrence graph.

\noindent
\textbf{Inductive Context Prediction Module}.
Let us first denote a general medical term as $t$. For a term-term co-occurrence graph, we treat all InV terms as possible context terms and denote them as $\{u_j\}_{j=1}^{|V|}$ where $|V|$ is the total number of terms in the graph. The inductive context prediction module aims to predict how likely term $u_j$ appears in the context of $t$ (denoted as the conditional probability $p\,(u_j|t)$). To learn a good context predictor, we utilize {all} existing terms in the graph as term $t$, i.e., $t \in \{u_i\}_{i=1}^{|V|}$ and the conditional probability becomes $p\,(u_j|u_i)$.

Formally, the probability of observing term $u_j$ in the context of term $u_i$ is denoted as:
\vspace{-10pt}
\begin{equation}
    p\,(u_j|u_i)= 
    \frac{\text{exp} \, (\nu_{u_j}^T\cdot s_{u_i})}
    {\sum_{k=1}^{|V|}\text{exp} \, (\nu_{u_k}^T \cdot s_{u_i})}
\end{equation}
where $s_{u_i}=h(u_i)$ and $h(\cdot)$ is the same encoder function defined in section \ref{subsubsec:bi-level-encoding}. $\nu_{u_j} \in \mathbb{R}^{d_o}$ is the context embedding vector corresponding to term $u_j$ and we let $d_o=d_s$. The predicted distribution $p\,(u_j|u_i)$ is optimized to be close to the empirical distribution $\hat{p}\,(u_j|u_i)$ defined as:
\vspace{-10pt}
\begin{equation}
    \hat{p}\,(u_j|u_i)= \frac{w_{ij}}{\sum_{(i,k)\in E} w_{ik}}
\end{equation}
where $E$ is the set of edges in the co-occurrence graph and $w_{ij}$ is the weight between term $u_i$ and term $u_j$. We adopt the cross entropy loss function for optimizing:
\begin{equation}
\label{eqn:context_loss}
    L_n= -\sum_{u_i,u_j \in V} \hat{p}(u_j|u_i)\ \text{log} \, (p(u_j|u_i)) 
\end{equation}

When the number of terms in the graph $|V|$ is very large, it is computationally costly to calculate the conditional probability $p\,(u_j|u_i)$, and one can utilize the negative sampling algorithm~\cite{mikolov2013efficient} to train our inductive context predictor efficiently. The loss function Eqn. \ref{eqn:context_loss} can be modified as:
\begin{equation}
    \log \sigma(\nu_{u_j}^T\cdot s_{u_i}) + \sum_{n=1}^{N_0} E_{u_n \sim P_n(u)}[\log \sigma (-\nu_{u_n}^T \cdot s_{u_i})]
\end{equation}
\noindent
where $\sigma(x)=1/(1+\exp(-x))$ and $u_n$ is the negative sample drawn from the noise distribution $P_n(u)\propto d_{u}^{3/4}$. $N_0$ is the number of negative samples and $d_{u}$ is the degree of term $u$ in the co-occurrence graph.

Now, given a term $t$ (either InV or OOV), we can select the top-$K$ terms as its predicted contexts based on the predicted probability distribution  $p\,(\cdot|t)$. Next, we describe the dynamic context matching mechanism to model the semantic similarity of two terms based on their predicted contexts.

\noindent
\textbf{Dynamic Context Matching {Mechanism}}.
Inspired by previous works on neighborhood aggregation based graph embedding methods~\cite{hamilton2017inductive, velickovic2017graph}, which generate an embedding vector for an InV node by aggregating features from its neighborhood (contexts), we introduce two semantic vectors respectively for the query term and the candidate term, $v_q, v_c \in \mathbb{R}^{d_e}$, and learn them by aggregating the feature vectors of their corresponding {top-$K$} predicted contexts from previous module.

Let us define $v_q^i \in \mathbb{R}^{d_e}$ as the feature vector of the $i$-th term in query term $q$'s context while $v_c^j \in \mathbb{R}^{d_e}$ as the feature vector of the $j$-th term in candidate term $c$'s context, and their context sets as $\Phi(q)=\{v_q^i\}_{i=1}^K$, $\Phi(c)=\{v_c^j\}_{j=1}^K$.
Essentially, as we aim to capture the semantic meaning of terms, the feature vectors $v_q^i$'s and $v_c^j$'s are expected to contain semantic information. Also noticing that all predicted context terms are InV terms (i.e., in the co-occurrence graph), which allows us to adopt widely used graph embeddings, such as LINE(2nd)~\cite{tang2015line} as their feature vectors.

One naive way to obtain the context semantic vectors, $v_q$ and $v_s$ is to average vectors in their respective context set. Since such $v_q$ (or $v_c$) does not depend on the other one, we refer to such vectors as "static" representations for terms.

\vspace{-10pt}
\begin{figure}[htbp!]
    \centering
    \includegraphics[width=0.9\linewidth]{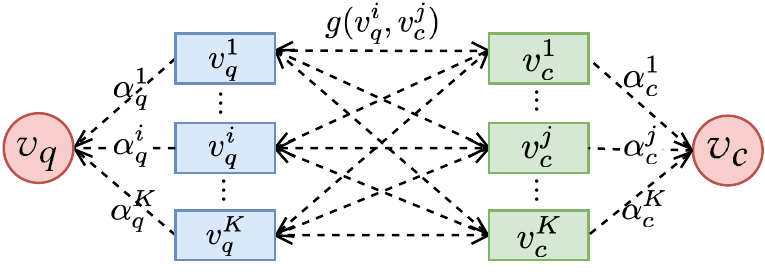}
    \vspace{-10pt}
    \caption{Dynamic Context Matching Mechanism.}
    \vspace{-10pt}
    \label{fig:dynamic_matching}
\end{figure}

In contrast to the static approach, we propose the \textit{dynamic context matching mechanism} (as shown in Figure \ref{fig:dynamic_matching}), which weighs each term in the context of $q$ (or $c$) based on its matching degree with terms in the context of $c$ (or $q$) and hence the context semantic vector representation $v_q$ (or $v_c$) is \textit{dynamically} changing depending on which terms it is comparing with. More specifically, let us define $g(x, y)=\text{tanh}(xW_my^T)$ as a nonlinear function parameterized with weight matrix $W_m\in \mathbb{R}^{d_e\times d_e}$ to measure the similarity between two row vectors $x$ and $y$. For each context vector $v_q^i$ of the query term, {we calculate its weight based on how it matches with $c$'s contexts overall}:
\begin{equation}
    \textsf{match} \,[v_q^i, \Phi(c)] = \textsf{Pooling}\, [g(v_q^i, v_c^1), ..., g(v_q^i, v_c^K)]
\end{equation}
For the pooling operation, we empirically choose the \textsf{mean} pooling strategy as it performs better than {alternatives such as \textsf{max} pooling} in our experiments. Then we normalize the weight of $v_q^i$ as:
\begin{equation}
    \alpha_q^i = \frac{\Large{\textit{e}}^{\;\textsf{match}[v_q^i, \Phi(c)]}}{\sum_{k=1}^K \Large{\textit{e}}^{\;\textsf{match}[v_q^{k}, \Phi(c)]}}
\end{equation}

Finally, the context semantic vector for the query term $v_q$ is calculated through a weighted combination of $q$'s contexts:
\begin{equation}
    v_q = \sum_{i=1}^K \alpha_q^i \cdot v_q^i
\end{equation}
 
Following the same procedure, we can obtain the context semantic vector $v_c$ for the candidate term w.r.t. the query term. Then we define the context score for a query term $q$ and a candidate term $c$ to measure their semantic similarity based on $v_q$ and $v_c$:
\begin{equation}
    \textsf{Context Score} \, (q, c)=f_c(v_q, v_c)
\end{equation}

\subsection{Model Optimization and Inference}
\label{subsec:train-inference}

\textbf{Objective Function.} 
Given a query term $q$ and a candidate term $c$, to capture their similarity based on surface forms and global contexts, we define the final score function as:
\begin{equation}
\label{eqn:final-score}
    f(q, c) = (1-\gamma) \cdot f_s(s_q, s_c) + \gamma \cdot f_c(v_q, v_c)
\end{equation}
\noindent
{$f_s(\cdot)$ and $f_c(\cdot)$ are similarity functions between two vectors, e.g., cosine similarity or bilinear similarity.}
Now we obtain the recommendation probability of each candidate $t_i \in \{t_1, ..., t_N\}$ given a query $q$:
\begin{equation}
    p(t_i|q)=\frac{\Large{\textit{e}}^{\, f(q, t_i)}}{\sum_{k=1}^{N} \Large{\textit{e}}^{\,f(q, t_k)}}
\end{equation}
where $N$ is the size of the candidate set. Finally, we adopt the ListNet~\cite{cao2007learning} ranking framework which minimizes the cross entropy loss for query term $q$:
\vspace{-10pt}
\begin{equation}
\label{eqn:ranking_loss}
    L_r= -\sum_{i=1}^{N} p^*(t_i|q) \ \text{log} \, p(t_i|q) 
\end{equation}
where $p^*(t_i|q)$ is the normalized ground-truth distribution of a list of ranking scores as $\{r_i\}_{i=1}^N$ where $r_i$ equals to $1$ if $q$ and $t_i$ are synonyms and $0$ otherwise.

\noindent 
\textbf{Training}.
For efficiency concerns, we adopt a two-phase training strategy: We first train the inductive context prediction module by loss function $L_n$ (Eqn. \ref{eqn:context_loss}) in the term-term co-occurrence graph, and sample top-K contexts based on the predicted probability distribution and use them in the context matching component. Then, we train the ranking framework by minimizing the ranking loss $L_r$ (Eqn. \ref{eqn:ranking_loss}). 

\noindent
\textbf{Inference}. At the inference stage, we treat all InV terms as candidates for a given query. Since the dynamic representation mechanism involves pairwise term matching between the contexts of the query term and those of each candidate term and can have a high computational cost when the candidate set size is large, we adopt a two-step strategy: (1) For a given query term, select its top-N high potential candidates based on the surface form encoding vector and the context semantic vector obtained by the static representation mechanism; (2) Re-rank the selected candidates by applying our \surfcon framework with the dynamic representation mechanism.

% \vspace{-10pt}
\section{Experiments}
\label{section:exp}

Now we evaluate our proposed framework \surfcon to show the effectiveness of leveraging both surface form information and global context information for synonym discovery.

\vspace{-10pt}
\subsection{Datasets}\label{exp:dataset}
\vspace{-2pt}

\noindent
\textbf{{Medical Term Co-occurence Graph.}} 
We adopt publicly available sets of medical terms with their co-occurrence statistics which are extracted by \citet{finlayson2014building} from 20 million clinical notes collected from Stanford Hospitals and Clinics\cite{lowe2009stride} since 1995. Medical terms are extracted using an existing phrase mining tool ~\cite{lependu2012annotation} by matching with 22 clinically relevant ontologies such as SNOMED-CT and MedDRA. And co-occurrence frequencies are counted based on how many times two terms co-occur in the same temporal \textit{bin} (i.e., a certain timeframe in patient's records), e.g., 1, 7, 30, 90, 180, 365, and $\infty$-day \textit{bins}. 

Without loss of generality, we choose 1-day per-bin and $\infty$-day per-bin\footnote{Per-bin means each unique co-occurring term-term pair is counted at most once for each relevant bin of a patient. We refer readers to \citet{finlayson2014building} for more information.} graphs to evaluate different methods. We first convert the global counts between nodes to the PPMI values \cite{levy2014linguistic} and adopt subsampling \cite{mikolov2013distributed} to filter very common terms, such as "medical history", "medication dose", etc. We choose these two datasets {because they have very different connection density as shown in Table \ref{tab:dataset-statistics}}, and denote them as {\textbf{1-day} and \textbf{All-day}} datasets.

\noindent
\textbf{Synonym Label.}
\label{synlabel} 
In the released datasets, \citet{finlayson2014building} provided a term-to-UMLS CUI mapping based on the same 22 ontologies as used when extracting terms. They reduced the ambiguity of  a term by suppressing its least likely meaning so as to provide a high-quality mapping. We utilized such mapping to obtain the synonym labels: Terms mapped to the same UMLS CUI are treated as synonyms, e.g., terms like "c vitamin", "vit c", "ascorbic acid" are synonyms as they are all mapped to the concept "Ascorbic Acid" with ID \text{C0003968}.

\noindent
\textbf{Query Terms.}
Given a medical term-term co-occurrence graph, terms in the graph that can be mapped to UMLS CUIs are treated as potential query terms, and we split all such terms into training, development and testing sets. Here, since all terms appear in the given co-occurrence graph, this testing set is referred to as the \textbf{InV testing set}. We also create an \textbf{OOV testing set}: Under a UMLS CUI, terms not in the co-occurrence graph are treated as OOV query terms and are paired with their synonyms which are in the graph to form positive pairs. We sample 2,000 of such OOV query terms for experiments. In addition, since synonyms with different surface forms tend to be more challenging to discover (e.g., "vitamin c" vs. "ascorbic acid"), we also sample a subset named \textbf{Dissim} under both \text{InV} and \text{OOV testing set}, where query terms paired with their dissimilar synonyms\footnote{Dissimilarity is measured by Levenshtein edit distance \cite{gomaa2013survey} with a threshold (0.8).} are selected. Statistics of our training/dev/testing sets are given in Table \ref{tab:dataset-statistics}. 

% \begin{table}[htbp!]
% % \vspace{-10pt}
% \caption{Statistics of our datasets.}
% % \vspace{-8pt}
% \resizebox{\linewidth}{!}{%
% \begin{tabular}{c|cccccc}
% \hline
% \multirow{2}{*}{} & \multirow{2}{*}{Train} & \multirow{2}{*}{Dev} & \multicolumn{2}{c}{InV Test} & \multicolumn{2}{c}{OOV Test} \\ 
% \cline{4-7} 
% & & & All & Dissim & All & Dissim \\ 
% \hline
% \multicolumn{7}{c}{1-day dataset}\\
% \hline
% \#Terms & 9,451 & 960 & 960 & 175 & 2,000 & 809 \\
% \hline
% \#Positive Pairs & 26,178 & 1,266 & 1,280 & 206 & 4,434 & 1,181 \\ 
% \hline
% \hline
% \multicolumn{7}{c}{All-day dataset}\\
% \hline
% \#Terms & 7,021 & 726 & 726 & 152 & 2,000 & 841 \\
% \hline
% \#Positive Pairs & 18,208 & 1,034 & 986 & 200 & 4,469 & 1,211 \\ 
% \hline
% \end{tabular}}
% % \begin{tabular}{c|cccc}
% % \hline
% % \multirow{2}{*}{} & \multirow{2}{*}{Train} & \multirow{2}{*}{Dev} & \multicolumn{2}{c}{Test} \\
% % \cline{4-5} 
% % & & & InV & OOV \\ 
% % \hline
% % \multicolumn{5}{c}{1-day dataset}\\
% % \hline
% % \#Terms & 9,451 & 960 & 960 & 2,000 \\
% % \hline
% % \#Positive Pairs & 26,179 & 1,267 & 1277 & 4,434 \\ 
% % \hline
% % \hline
% % \multicolumn{5}{c}{All-day dataset}\\
% % \hline
% % \#Terms & 7,021 & 726 & 726 & 2000 \\
% % \hline
% % \#Positive Pairs & 18,184 & 1,031 & 987 & 4,469\\ 
% % \hline
% % \end{tabular}
% \label{tab:dataset-statistics}
% \end{table}
% % \vspace{-10pt}
\begin{table}[htbp!]
\centering
\vspace{-5pt}
\caption{Statistics of our datasets.}
\vspace{-5pt}
% \resizebox{\linewidth}{!}{%
\nop{
\begin{tabular}{|c|l|cc|}
\hline
\multicolumn{2}{|c|}{}                      & \textbf{1-day} dataset & \textbf{All-day} dataset \\ \hline
\multicolumn{2}{|c|}{\# Nodes}              & 52,804        & 43,406          \\
\multicolumn{2}{|c|}{\# Edges}              & 16,197,319    & 50,134,332      \\
\multicolumn{2}{|c|}{Average \# Degrees}    & 613.5         & 2310.0          \\ 
\hline
\hline
% \hline
\multicolumn{2}{|c|}{\# Train Term}        & 9,451         & 7,021           \\
\multicolumn{2}{|c|}{\# Dev Term}          & 960           & 726             \\
\multicolumn{2}{|c|}{\# InV Test Tm (All)}     & 960           & 726             \\
\multicolumn{2}{|c|}{\# InV Terms (Dissim)} & 175           & 152             \\
\multicolumn{2}{|c|}{\# OOV Test Term (All)}    & 2,000         & 2,000           \\
\multicolumn{2}{|c|}{\# OOV Terms (Dissim)} & 809           & 841             \\ 
\hline
\end{tabular}
}

    \begin{tabular}{c|c|cc}
    \hline
    \multicolumn{2}{c|}{}         & 1-day dataset & All-day dataset \\ 
    \hline
    \multicolumn{2}{c|}{\# Nodes} & 52,804 & 43,406   \\
    \multicolumn{2}{c|}{\# Edges} & 16,197,319 & 50,134,332  \\
    \multicolumn{2}{c|}{Average \# Degrees} & 613.5 & 2310.0\\ 
    \hline
    \hline
    \multicolumn{2}{c|}{\# Train Terms} & 9,451 & 7,021\\
    \multicolumn{2}{c|}{\# Dev Terms} & 960 & 726 \\ 
    \hline
    \multirow{2}{*}{\# InV Test Terms} & \text{All} & 960 & 726 \\
                              & \text{Dissim} & 175 & 152 \\ 
                              \hline
    \multirow{2}{*}{\# OOV Test Terms} & \text{All}    & 2,000 & 2,000 \\
                              & \text{Dissim} & 809 & 841 \\ 
                              \hline
                              
    \end{tabular}
    \vspace{-20pt}
    \label{tab:dataset-statistics}
\end{table}

\begin{table*}[htbp!]
% \vspace{-5pt}
\caption{Model evaluation {in \textsf{MAP}} with {random candidate selection}.
\vspace{-5pt}
\nop{Names are not good. How about "Using surface form info only", "using  global contexts only", "using both"..."semantic" is too vague..} 
\nop{I think you may just test "all" and "Dissim" and do not need to test Sim. "all" means without removing lexical variants and "Dissim" means removing lexical variants from the candidate term list of each query.}}
\resizebox{\linewidth}{!}{%
\begin{tabular}{c|c|ccccc|ccccc}
\hline
\multirow{3}{*}{Method Category} & \multirow{3}{*}{Methods} & \multicolumn{5}{c|}{\textbf{1-day} Dataset} & \multicolumn{5}{c}{\textbf{All-day} Dataset}  \\ 
\cline{3-12} 
& & \multicolumn{1}{c}{\multirow{2}{*}{Dev}} & \multicolumn{2}{c}{\text{InV Test}} & \multicolumn{2}{c|}{\text{OOV Test}} 
& \multicolumn{1}{c}{\multirow{2}{*}{Dev}} & \multicolumn{2}{c}{\text{InV Test}} & \multicolumn{2}{c}{\text{OOV Test}} \\ 
\cline{4-7} \cline{9-12}
& & \multicolumn{1}{c}{} & \text{All} & \multicolumn{1}{c}{\text{Dissim}} & \text{All} & \text{Dissim} & \multicolumn{1}{c}{} & \text{All} & \multicolumn{1}{c}{\text{Dissim}} & \text{All} & \text{Dissim} 
\\ 
\hline
% \multicolumn{11}{c}{String-based methods}\\
%\hline
\multirow{3}{*}{\shortstack[c]{\begin{tabular}[c]{@{}c@{}}Surface form \\ based methods\end{tabular}\nop{ Using surface \\ form info only}}} 
% String Similarity 
% & 0.7976 & 0.7731 & 0.1836 & 0.6161 & 0.1752 & 0.7421 & 0.7399 & 0.2050 & 0.6036 & 0.1769 \\
& CharNgram~\cite{hashimoto2017jmt}
& 0.8755 & 0.8473 & 0.4657 & 0.7427 & 0.4131 & 0.8652 & 0.8553 & 0.4615 & 0.7675 & 0.4424 \\ 
& CHARAGRAM~\cite{wieting2016charagram}
& 0.8705 & 0.8507 & 0.5504 & 0.7609 & 0.5142 & 0.8915 & 0.8805 & 0.5153 & 0.8119 & 0.5282 \\ 
% & MaLSTM~\cite{mueller2016siamese}
% & 0.7478 & 0.7246 & 0.4693 & 0.5742 & 0.3161 & 0.7239 & 0.6864 & 0.3943 & 0.5392 & 0.3180 \\ 
& SRN~\cite{neculoiu2016learning}
& 0.8886 & 0.8565 & 0.5102 & 0.7241 & 0.4341 & 0.8460 & 0.8170 & 0.4523 & 0.7110 & 0.4176 \\
% Word2vec~\cite{mikolov2013distributed} 
% & 0.36485 & 0.37904 & 0.30831 & - & - & 0.37188 & 0.38359 & 0.36448 & - & - \\ 
% LINE2~\cite{tang2015line}
% & 0.40351 & 0.41712 & 0.35142 & - & - & 0.38019 & 0.38955 & 0.36918 & - & - \\ 
% CHARNGRAM~\cite{hashimoto2017jmt}
% & 0.84384 & 0.8132 & 0.41047 & 0.65505 & 0.33892 & 0.79054 & 0.78234 & 0.34871 & 0.65686 & 0.34325 \\
% \hline
% \hline
\hline\hline
\multirow{3}{*}{\shortstack[c]{\begin{tabular}[c]{@{}c@{}}Global context \\ based methods\end{tabular}\nop{ Using global \\ contexts only}}} & 
% \multicolumn{11}{c}{Semantic-based methods}\\
Word2vec~\cite{mikolov2013distributed}
& 0.3838 & 0.3748 & 0.3188 & - & - & 0.4801 & 0.476 & 0.4180 & - & - \\ 
% DPE-NoP~\cite{qu2017automatic} 
& LINE(2nd)~\cite{tang2015line}
& 0.4279 & 0.4301 & 0.3494 & - & - & 0.5068 & 0.5043 & 0.4369 & - & - \\ 
& DPE-NoP~\cite{qu2017automatic}
& 0.6222 & 0.6107 & 0.4855 & - & - & 0.5928 & 0.5949 & 0.4938 & - & - \\ 
\hline\hline
\multirow{2}{*}{\shortstack[c]{
\begin{tabular}[c]{@{}c@{}}Hybrid methods \\  (surface+context)\end{tabular}\nop{Using both}}} & 
% \multicolumn{11}{c}{Hybrid methods}\\
Concept Space~\cite{wang2015medical}
& 0.8094 & 0.8109 & 0.4690 & - & - & 0.8064 & 0.7924 & 0.5574 & - & - \\ 
& Planetoid~\cite{yang2016revisiting} 
& 0.8813 & 0.8514 & 0.5612 & 0.731 & 0.4714 & 0.8818 & 0.8765  & 0.6963 & 0.7403 & 0.4986 \\
% & Planetoid~\cite{yang2016revisiting} 
% & 0.9078 & 0.89205 & 0.66803 & 0.77719 & 0.52687 & 0.8818 & 0.8765  & 0.6963 & 0.7403 & 0.4986 \\
\hline
\hline
\multirow{3}{*}{\shortstack[c]{\begin{tabular}[c]{@{}c@{}}Our model \\ and variants\end{tabular}}} & 
% \multicolumn{11}{c}{Our framework and its variants:}\\
\text{SurfCon (Surf-Only)}
& 0.9160 & 0.9053 & 0.6145 & 0.8228 & 0.5829 & 0.9034 & 0.8958 & 0.6006 & 0.8183 & 0.5622 \\
% &SurfCon (Con-Only)
% & 0.6559 & 0.6470 & 0.5092 & 0.4909 & 0.3595 & 0.0000 & 0.0000 & 0.0000 & 0.0000 & 0.0000 \\
% & \text{SurfCon (LINE2)} & 0.9187 & 0.9044 & 0.6199 & - & - & 0.9199 & 0.9135 & 0.7402 & - & - \\
& \text{SurfCon (Static)}
& 0.9242 & 0.9151 & 0.6542 & 0.8285 & 0.5933 & 0.9170 & 0.9019 & 0.6656 & 0.8203 & 0.5664 \\
& \textbf{SurfCon}
& \textbf{0.9348} & \textbf{0.9176} & \textbf{0.6821} & \textbf{0.8301} & \textbf{0.6009} & \textbf{0.9219} & \textbf{0.9199} & \textbf{0.7171} & \textbf{0.8232} & \textbf{0.5673} \\
\hline
\end{tabular}}
\vspace{-10pt}
\label{tab:main-results}
\end{table*}

\vspace{-8pt}
\subsection{Experimental Setup}
\label{exp:setup}
\vspace{-2pt}

\subsubsection{Baseline methods.}
\label{baseline}
We compare \surfcon with the following 10 methods. {The baselines} can be categorized by three types: (i) Surface form based methods, which focus on capturing the surface form information of terms. (ii) Global context based methods, which try to learn embeddings of terms for synonym discovery; (iii) Hybrid methods, which combine surface form and global context information. The others are our model variants.

\noindent
\textbf{Surface form based methods}.
    (1)
    \textit{CharNgram}~\cite{hashimoto2017jmt}:
    We borrow pre-trained character n-gram embeddings from ~\citet{hashimoto2017jmt} and take the average of unique n-gram embeddings for each term as its feature, and then train a bilinear scoring function following previous works~\cite{qu2017automatic, zhang2019synonymnet}.
    (2)
    \textit{CHARAGRAM} \cite{wieting2016charagram}:
    Similar as above, but we further fine-tune CharNgram embeddings using synonym supervision.
    (3)
    \textit{SRN} \cite{neculoiu2016learning}:
    A Siamese network structure is adopted with a bi-directional LSTM to encode character sequence of each term and cosine similarity is used as the scoring function.
    
\noindent
\textbf{Global context based methods}.
    (4) \textit{Word2vec} \cite{mikolov2013distributed}:
    A popular distributional embedding method. We obtain word2vec embeddings by doing {SVD decomposition over the Shifted PPMI co-occurrence matrix \cite{levy2014neural}}. We treat the embeddings as features and use a bilinear score function for synonym discovery.
    (5) \textit{LINE(2nd)} \cite{tang2015line}:
    A widely-adopted graph embedding approach. Similarly, embeddings are treated as features and a bilinear score function is trained to detect synonyms.
    (6) \textit{DPE-NoP} \cite{qu2017automatic}:
    DPE is proposed for synonym discovery on text corpus, and consists of a distributional module and a pattern module, where the former utilizes global context information and the latter learns patterns from raw sentences. Since raw texts are unavailable in our setting, we only deploy the distributional module (a.k.a. DPE-NoP in \citet{qu2017automatic}).
    
\noindent
\textbf{Hybrid methods}.
    (7)
    \textit{Concept Space Model} \cite{wang2015medical}: 
    A medical synonym extraction method that combines word embeddings and heuristic rule-based string features.
    (8)
    \textit{Planetoid} \cite{yang2016revisiting}: 
    An inductive graph embedding method that can generate embeddings for both observed and unseen nodes. We use the bi-level surface form encoding vectors as the input and take the intermediate hidden layer as embeddings. Similarly, a bilinear score function is used for synonym discovery.
    
\noindent
\textbf{Model variants}.
    (9)
    \textit{\surfcon (Surf-Only)}:
    A variant of our framework which only uses the surface score for ranking.
    (10)
    \textit{\surfcon (Static)}:
    Our framework with static representation mechanism. By comparing these variants, we verify the performance gain brought by modeling global contexts using different matching mechanisms. 

For baseline methods (1-3 and 8) and our models, we test them under both InV and OOV settings. For the others (4-7), because they rely on embeddings that are only available for InV terms, we only test them under InV setting. 

\vspace{-5pt}
\subsubsection{Candidate Selection and Performance Evaluation.}
For evaluating baseline methods and our model, we experiment with two strategies: 
(1) Random candidate selection. For each query term, we randomly sample 100 non-synonyms as negative samples and mix them with synonyms for testing. This strategy is widely adopted by previous work on synonym discovery for testing efficiency~\cite{wang2015medical, zhang2019synonymnet}.
(2) Inference-stage candidate selection.
As mentioned in section \ref{subsec:train-inference}, at the inference stage, we first obtain high potential candidates in a lightweight way. Specifically, after the context predictor is pre-trained, for all terms in the given graph as well as the query term, we generate their surface form vector $s$ and context semantic vector $v$ obtained by the static representation. Then we find top 50 nearest neighbors of the query term respectively {based on} $s$ and $v$ using cosine similarity. Finally, we apply {our methods and baselines} to re-rank the 100 high potential candidates. {We refer to these two strategies as \textit{random candidate selection} and \textit{inference-stage candidate selection}.}

For evaluation, we adopt a popular ranking metric Mean Average Precision defined as $\textsf{MAP}=\frac{1}{|Q|} \sum_{i=1}^{|Q|}\frac{1}{m_i} \sum_{j=1}^{m_i} \textsf{Precision}(R_{ij})$, where $R_{ij}$ is the set of ranked terms from $1$ to $j$, $m_i$ is the length of $i$-th list, and $|Q|$ is the number of queries.

\vspace{-5pt}
\subsubsection{Implementation details}
\label{details}
Our framework is implemented in Pytorch \cite{paszke2017automatic} with Adam optimizer \cite{kingma2014adam}.
The dimensions of character embeddings ($d_c$), word embeddings ($d_w$), surface vectors ($d_s$), and sementic vectors ($d_e$) are set to be 100, 100, 128, 128. Early stopping is used when the performance in the dev sets does not increase continuously for 10 epochs.
We directly optimize Eqn. \ref{eqn:context_loss} since the number of terms in our corpus is not very large, and set $f_s(\cdot)$ and $f_c(\cdot)$ to be cosine similarity and bilinear similarity function respectively, based on the model performance on the dev sets.
When needed, string similarities are calculated by using the Distance package\footnote{https://github.com/doukremt/distance}. 
Pre-trained CharNgram \cite{hashimoto2017jmt} embeddings are borrowed from the authors\footnote{https://github.com/hassyGo/charNgram2vec}. 
For CHARAGRAM \cite{wieting2016charagram}, we initialize the n-gram embeddings by using pre-trained CharNgram and fine-tune them on our dataset by the synonym supervision. 
We learn LINE(2nd) embeddings \cite{tang2015line} by using OpenNE\footnote{https://github.com/thunlp/OpenNE}.
Heuristic rule-based matching features of Concept Space model are implemented according to ~\cite{wang2015medical}.
Code, datasets, and more implementation details are available online\footnote{\url{https://github.com/yzabc007/SurfCon}}.

\vspace{-5pt}
\subsection{Results and Analysis}
\label{main-results}

\subsubsection{Evaluation with {Random Candidate Selection}}
We compare all methods under random candidate selection strategy with the results shown in Table \ref{tab:main-results}.

\noindent
\textbf{(1) Comparing \surfcon with surface form based methods.} \\
Our model beats all surface form based methods, including strong baselines such as SRN that use complicated sequence models to capture character-level information. This is because: 1) Bi-level encoder of \surfcon could capture surface form information from both character- and word-level, while baselines only consider either of them; 2) \surfcon captures global context information, which could complement surface form information for synonym discovery. In addition, in comparison with CharNgram and CHARAGRAM, our model variant \surfcon (Surf-Only), which also only uses surface form information, obtains consistently better performance, especially in the OOV Test set. The results demonstrate that adding word-level surface form information is useful to discover synonyms.

\noindent
\textbf{(2) Comparing \surfcon with global context based methods.} \\
\surfcon substantially outperforms all other global context based methods (Word2vec, LINE(2nd) and DPE-NoP). This is largely due to the usage of surface form information. In fact, as one can see, global context based methods are generally inferior to surface form based methods, partly due to the fact that a large part of synonyms are similar in surface form, while only a small portion of them are in very different surface form. Thus, detecting synonyms without leveraging surface information can hardly lead to good results. Besides, our context matching component conducts context prediction and matching strategies, which takes better advantage of global context information and thus lead to better performance on the synonym discovery task.

\noindent
\textbf{(3) Comparing SurfCon with hybrid methods.}
We also compare our model with baselines that combine both surface form and global context information. First, \surfcon is superior to the concept space model because the latter simply concatenates distributional embeddings with rule-based string features, e.g., the number of shared words as features and apply a logistic regression classifier for classification. Further, \surfcon also performs better than Planetoid, partly because our framework more explicitly leverages both surface form and global context information to formulate synonym scores, while Planetoid relies on one embedding vector for each term which only uses surface form information as input.

\noindent
\textbf{(4) Comparing \surfcon with its variants.}
To better understand why \surfcon works well, we compare it with several variants. Under {both datasets}, \surfcon (Surf-Only) {already outperforms} all baselines demonstrating the effectiveness of our bi-level surface form encoding component. With the context matching component in \surfcon (Static), the performance is further improved, especially under \textit{InV Test Dissim} setting where synonyms tend to have different surface forms and we observe around 4\% performance gain. Further, by using dynamic representation in context matching mechanism, \surfcon obtains better results, which demonstrates that the dynamic representation is more effective to utilize context information compared with the static strategy.

\begin{table}[t!]
\centering
\caption{Model evaluation at inference stage.}
\vspace{-5pt}
\resizebox{.45\textwidth}{!}{%
\begin{tabular}{c|cccc}
\hline
\multirow{2}{*}{Methods} & \multicolumn{2}{c}{1-day} & \multicolumn{2}{c}{All-day} \\ 
\cline{2-5}
& InV Test & OOV Test & InV Test  & OOV Test \\ 
\hline
CHARAGRAM~\cite{hashimoto2017jmt}  & 0.3921 & 0.4044  & 0.3941 & 0.3913 \\
DPE-NoP~\cite{qu2017automatic}  & 0.2396 & -  & 0.2408 & - \\
Planetoid~\cite{yang2016revisiting} & 0.4563 & 0.4268  & 0.3765 & 0.3812 \\ 
\surfcon  & 0.5525 & 0.5068 & 0.4686 & 0.4661 \\
\hline
\end{tabular}}
\vspace{-12pt}
\label{main-results-practical}
\end{table}
\vspace{-5pt}
\subsubsection{Evaluation at Inference Stage}
To further evaluate the power of our model in real practice, we test its performance at the inference stage as mentioned in section \ref{subsec:train-inference}. Due to space constraint, we only show the comparison in Table \ref{main-results-practical} between \surfcon and several strong baselines revealed by Table \ref{tab:main-results}. In general, the  performance of all methods decreases at the inference stage compared with the random candidate selection setting, because the constructed list of candidates becomes harder to rank since surface form and context information are already used for the construction. For example, a lot of non-synonyms with similar surface form are often included in the candidate list. Even though the task becomes harder, we still observe our model outperforms the strong baselines by a large margin (e.g., around 8\% at least) under all settings.

\begin{figure}[t!]
    \centering
    \resizebox{\linewidth}{!}{
    \subfloat{\includegraphics[]{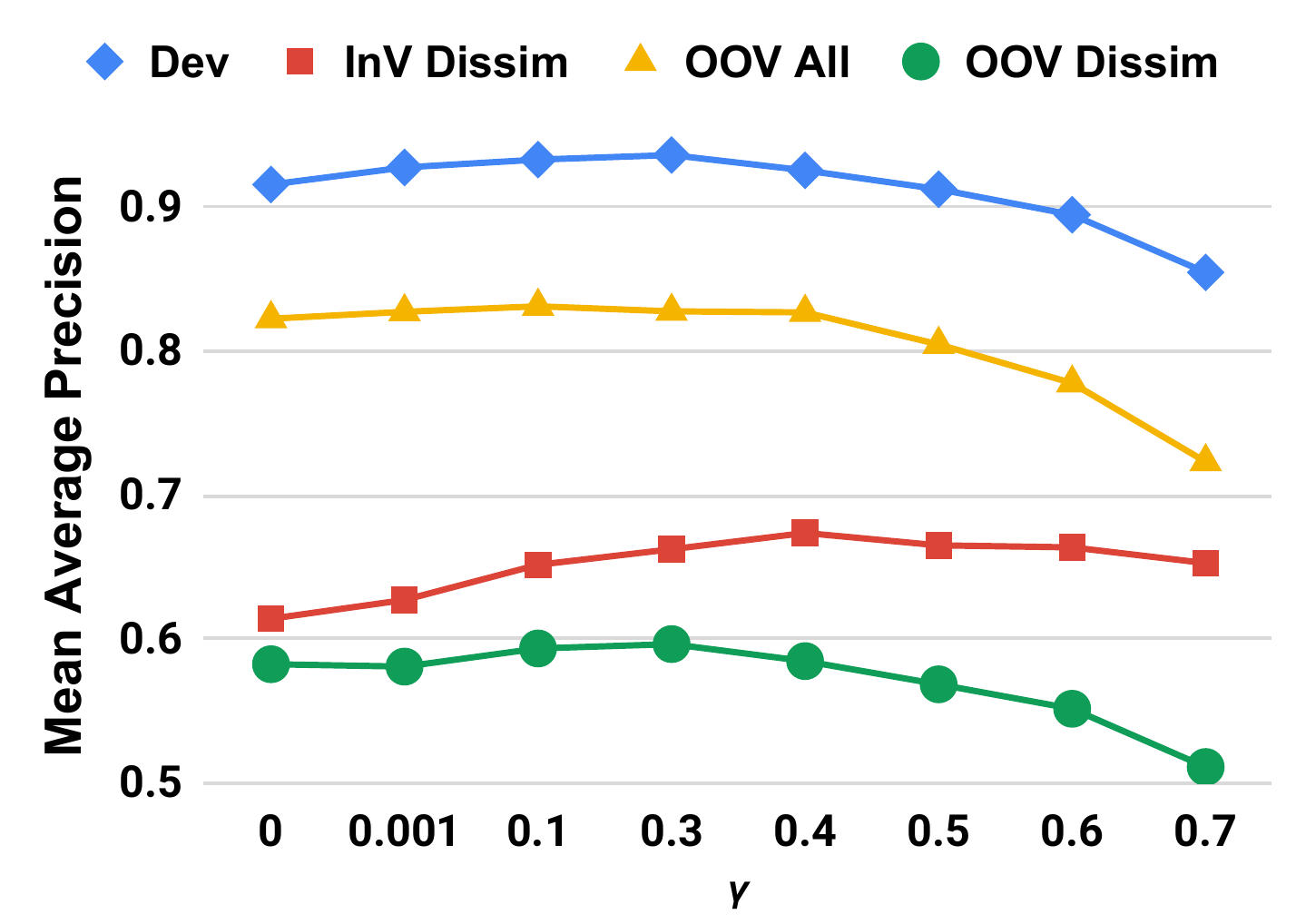}}  
    \subfloat{\includegraphics[]{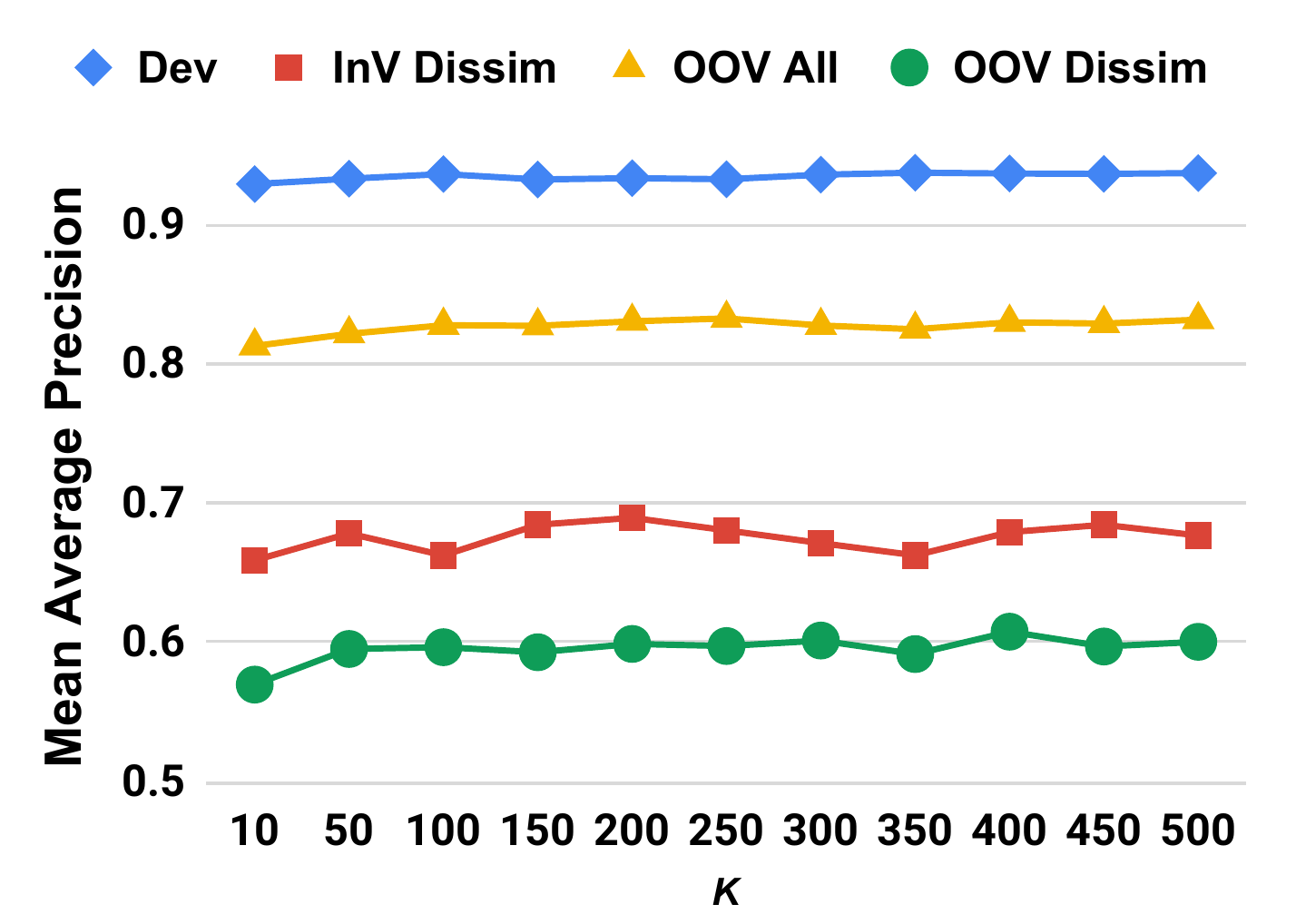}}    
    % \begin{subfigure}[t]{0.25\textwidth}
    % \centering
    % \includegraphics[height=1.23in,left]{para_gamma_final.pd}
    % \vspace{-12pt}
    % \caption{Coefficient of context score}
    % \end{subfigure}
    ~
    % \begin{subfigure}[t]{0.25\textwidth}
    % \centering
    % \includegraphics[height=1.23in,right]{para_num_contexts_final.pdf}
    % \vspace{-12pt}
    % \caption{Number of contexts}
    % \end{subfigure}
    }
    \vspace{-10pt}
    \caption{{Performance w.r.t. (a) the coefficient of context score $\gamma$  and (b) the number of context terms $K$}.}
    \label{fig:parameter_sensitivity}
    \vspace{-16pt}
\end{figure}

\vspace{-5pt}
\subsubsection{Parameter Sensitivity}
Here we investigate the effect of two important hyper-parameters: The coefficient $\gamma$ which balances the surface score and the context score, and the number of predicted contexts $K$ used for context matching. As shown in Figure \ref{fig:parameter_sensitivity}(a), the performance of \surfcon first is improved as $\gamma$ increases, which is expected because as more semantic information is incorporated, \surfcon could detect more synonyms that are semantically similar. When we continue to increase $\gamma$, the performance begins to decrease and the reason is that surface form is also an important source of information that needs to be considered. \surfcon achieves the best performance roughly at $\gamma=0.3$ indicating surface form information is relatively more helpful for the task than global context information. This also aligns well with our observation that synonyms more often than not have similar surface forms. Next, we show the impact of $K$ in Figure \ref{fig:parameter_sensitivity}(b). In general, when $K$ is small (e.g., $K=10$), the performance is not as good since little global context information is considered. Once $K$ increases to be large enough (e.g., $\geq50$), the performance is not sensitive to the variation under most settings showing that we can choose smaller $K$ for computation efficiency but still with good performance.

\vspace{-8pt}
\begin{table}[hbtp!]
\centering
\caption{Case studies on the 1-day dataset. {Bold terms are synonyms in our labeled set while underlined terms are not but quite similar to the query term in semantics.}}
\vspace{-8pt}
\resizebox{.4\textwidth}{!}{
\begin{tabular}{c|c|c}
\hline
\textsf{Query Term}                                                                        & \begin{tabular}[c]{@{}c@{}}"unable to vocalize" \\ (InV)\end{tabular} & \begin{tabular}[c]{@{}c@{}}"marijuana" \\ (OOV)\end{tabular} \\ 
\hline
\multirow{5}{*}{\begin{tabular}[c]{@{}c@{}}\textsf{\surfcon}\\\textsf{Top Ranked} \\ \textsf{Candidates} \end{tabular}} & \ul{"does not vocalize"}                                                 & \textbf{"marijuana abuse"}                                            \\
                                                                                  & \ul{"aphonia"}                                                             & \textbf{"cannabis"}                                                   \\
                                                                                  & \ul{"loss of voice"}                                                      &  \ul{"cannabis use"}                                                   \\
                                                                                  & "vocalization"                                                        & \ul{"marijuana smoking"}                                          \\
                                                                                  & \textbf{"unable to phonate"}                                                  & "narcotic"                                             \\ 
\hline

\multirow{3}{*}{\begin{tabular}[c]{@{}c@{}}\textsf{Labeled} \\ \textsf{Synonym}\\\textsf{Set}\end{tabular}}
& \multirow{3}{*}{"unable to phonate"}                                  & "cannabis"                                                   \\
                                                                                  &                                                                     & "marijuana abuse"                                        \\  
                                                                                  &                                                                     & "marihuana abuse"                                            \\ \hline
\end{tabular}}
\vspace{-10pt}
\label{tab:case_study}
\end{table}
\vspace{-8pt}
\subsection{Case Studies}
\vspace{-2pt}
We further conduct case studies to show the effectiveness of \surfcon. Two query terms {"unable to vocalize"} and {"marijuana"} are chosen respectively from the InV and OOV test set where the former is defined as the inability to produce voiced sound and the latter is a psychoactive drug used for medical or recreational purposes. As shown in Table \ref{tab:case_study}, for the InV query {"unable to vocalize"}, our model can successfully detect its synonyms such as "unable to phonate", which already exists in the labeled synonym set collected based on term-to-UMLS CUI mapping as we discussed in Section \ref{task-setting}. More impressively, our framework also discovers some highly semantically similar terms such as "does not vocalize" and "aphonia", even if some of them are quite different in surface form from the query term. For the OOV query {"marijuana"}, \surfcon ranks its synonym "marijuana abuse" and "cannabis" at a higher place. Note that the other top-ranked terms are also very relevant to "marijuana". 

\vspace{-8pt}
\section{Related Work}
\vspace{-2pt}
\noindent 
\textbf{Character Sequence Encoding.} To capture the character-level information of terms, neural network models such as Recurrent Neural Networks and Convolutional Neural Networks can be applied on character sequences ~\cite{ballesteros2015improved,kim2016character}. Further, CHARAGRAM~\cite{wieting2016charagram}, FastText~\cite{bojanowski2016enriching}, and CharNGram~\cite{hashimoto2017jmt} are proposed to represent terms and their morphological variants by capturing the shared subwords and $n$-grams information. However, modeling character-level sequence information only is less capable of discovering semantically similar synonyms,
{and our framework considers global context information to discover those synonyms.}

\noindent 
\textbf{Word and Graph/Network Embedding.} 
Word embedding methods such as word2vec~\cite{mikolov2013distributed} and Glove ~\cite{pennington2014glove} have been proposed and successfully applied to mining relations of medical phrases~\cite{wang2015medical,pakhomov2016corpus}. More recently, there has been a surge of graph embedding methods that seek to encode structural graph information into low-dimensional dense vectors, such as Deepwalk~\cite{perozzi2014deepwalk}, LINE~\cite{tang2015line}. Most of the embedding methods can only learn embedding vectors for words in the corpus or nodes in the graph, and thus fail to address the OOV issue. On the other hand, some more recent inductive graph embedding works, such as Planetoid~\cite{yang2016revisiting}, GraphSAGE ~\cite{hamilton2017inductive}, and SEANO~\cite{liang2018semi}, could generate embeddings for nodes that are unobserved in the training phase by utilizing their node features (e.g., text attributes). \textit{However, most of them assume the neighborhood of those unseen nodes is known, which is not the case for our OOV issue as the real contexts of an OOV term are unknown.} Since Planetoid~\cite{yang2016revisiting} can generate node embeddings based on node features such as character sequence encoding vectors, it can handle the OOV issue and is chosen as a baseline model.

\noindent 
\textbf{Synonym Discovery.} 
A variety of methods have been proposed to detect synonyms of medical terms, ranging from utilizing lexical patterns~\cite{weeds2004characterising} and clustering~\cite{matsuo2006graph} to the distributional semantics models~\cite{hagiwara2009supervised}. There are some more recent works on automatic synonym discovery ~\cite{wang2015medical,qu2017automatic,zhang2019synonymnet, Shen2019SynSetMine}. For example, \citet{wang2015medical} try to learn better embeddings for terms in medical corpora by incorporating their semantic types and then build a linear classifier to decide whether a pair of medical terms is synonyms or not. \citet{qu2017automatic} combine distributional and pattern based methods for automatic synonym discovery. However, many aforementioned models focus on finding synonyms based on raw texts information, which is not suitable for our privacy-aware clinical data. In addition, nearly all methods could only find synonyms for terms that appear in the training corpus and, thus cannot address the OOV query terms.

\vspace{-8pt}
\section{Conclusion}
\vspace{-2pt}
In this paper, we study synonym discovery on privacy-aware clinical data, which is a new yet practical setting and consumes less sensitive information to discover synonyms. We propose a novel and effective framework named \surfcon that considers both the surface form information and the global context information, can handle both InV and OOV query terms, and substantially outperforms various baselines on real-world datasets. As future work, we will extend \surfcon to infer more semantic relationships (besides synonymity) between terms and test it on more real-life datasets.

%
% The acknowledgments section is defined using the "acks" environment (and NOT an unnumbered section). This ensures
% the proper identification of the section in the article metadata, and the consistent spelling of the heading.
\vspace{-8pt}
\begin{acks}
\vspace{-2pt}
This research was sponsored in part by the Patient-Centered Outcomes Research Institute Funding ME-2017C1-6413, the Army Research Office under cooperative agreements W911NF-17-1-0412, NSF Grant IIS1815674, and Ohio Supercomputer Center \cite{OhioSupercomputerCenter1987}. The views and conclusions contained herein are those of the authors and should not be interpreted as representing the official policies, either expressed or implied, of the Army Research Office or the U.S.Government. The U.S. Government is authorized to reproduce and distribute reprints for Government purposes notwithstanding any copyright notice herein.
\end{acks}

% \bibliographystyle{ACM-Reference-Format}
% \bibliography{kdd}

\end{document}